%% file: main.tex
\documentclass[journal]{IEEEtran}
\usepackage{amsmath,amsfonts}
\usepackage{array}
\usepackage{url}
\usepackage{graphicx}
\usepackage{booktabs}
\usepackage{nicefrac}
\usepackage{microtype}
\usepackage{xcolor}
\usepackage{makecell}
\usepackage{multirow}
\usepackage{xspace}
\usepackage{wrapfig}
\usepackage{color, colortbl}
\usepackage{float}
\usepackage{cite}
\usepackage[absolute,overlay]{textpos}
\usepackage{hyperref}

\providecommand{\citep}[1]{\cite{#1}}
\providecommand{\citet}[1]{\cite{#1}}

\let\appendix\appendices

\def\ie{\emph{i.e., }}
\def\eg{\emph{e.g., }}

\title{Unified Reward Model for Multimodal Understanding and Generation}
\author{
Yibin Wang$^{1,2}$ \quad
Yuhang Zang$^{3\dagger}$ \quad
Hao Li$^{1,2}$ \quad
Cheng Jin$^{1,2\dagger}$ \quad
Jiaqi Wang$^{2,3\dagger}$ \\[6pt]
{$^{1}$ Fudan University \quad $^{2}$ Shanghai Innovation Institute \quad $^{3}$ Shanghai AI Lab} \\
Project Page: \href{https://codegoat24.github.io/UnifiedReward/}{codegoat24.github.io/UnifiedReward}
}

\newcommand{\ourmethod}{{\textsc {UnifiedReward}}\xspace}

\begin{document}

\maketitle
\begingroup
\renewcommand{\thefootnote}{\fnsymbol{footnote}}
\footnotetext[2]{Corresponding authors.}
\endgroup

\input{0.abstract}

\input{1.intro}

\input{2.related_work}

\input{3.method}

\input{4.exp}

\input{5.conclusion}

{
    \small
    \bibliographystyle{IEEEtran}
    \bibliography{main}
}

\input{6.appendix}


\end{document}

%% file: 0.abstract.tex
\begin{abstract}
Recent advances in human preference alignment have significantly improved multimodal generation and understanding. A key approach is to train reward models that provide supervision signals for preference optimization. However, existing reward models are often task-specific, limiting their adaptability across diverse visual applications. We also argue that a reward model that jointly learning to assess multiple vision tasks may foster a synergistic effect, where improved image understanding enhances image generation assessment, and refined image evaluation benefits video assessment through better frame analysis.
To this end, this paper proposes \ourmethod, the first unified reward model for multimodal understanding and generation assessment. It supports both pairwise ranking and pointwise scoring, providing effective reward signals for vision model preference alignment. Specifically, (1) we first train \ourmethod on our constructed large-scale human preference dataset, which covers both image and video generation/understanding tasks. (2) Then, we leverage it to automatically construct high-quality pairwise preference data from vision models by progressively filtering their outputs through our two-stage strategy, \ie pair ranking and point sifting.
(3) Finally, we use these data to align vision models with human preferences via Direct Preference Optimization (DPO). Experimental results show that jointly learning to assess diverse visual tasks yields substantial mutual benefits. We further apply our pipeline to both vision understanding and generation, achieving consistent improvements across each domain.
\end{abstract}

%% file: 1.intro.tex
\section{Introduction}
\IEEEPARstart{R}{ecent} advancements in human preference alignment have substantially propelled the progress of multimodal generation and understanding tasks. A straightforward technique is to directly collect human feedback to construct preference datasets for model optimization \citep{diffusiondpo,videodpo,llava-houd-dpo}. Despite its effectiveness, collecting large-scale human feedback is time-consuming and resource-intensive. To this end, an alternative popular approach involves learning reward models \citep{LiFT,llava-critic,zang2025internlm,videoreward,lee2023aligning,li2025temporal} from a limited amount of preference data and using the learned reward function to generate preference data based on the output of vision models. This synthetic preference data can then be leveraged for vision model preference alignment, significantly reducing the need for extensive human annotations.

\input{figs/shortcut1}

Despite their progress, we posit two concerns: \textbf{(1)} current reward models are often tailored to specific tasks, as shown in Tab. \ref{tab:reward_model_comparison}, limiting their adaptability across diverse visual understanding and generative tasks.
The key challenge lies in the lack of a comprehensive human preference dataset that spans a wide range of visual tasks. \textbf{(2)} We intuitively argue that visual tasks are inherently interconnected, and jointly learning multiple visual tasks may create a mutually reinforcing effect. Specifically, enhanced image understanding may improve the evaluation of image generation by providing a more accurate assessment of content quality and contextual relevance. Similarly, improvements in image evaluation may benefit video evaluation, as high-quality image assessments lead to more accurate evaluations of video frames, contributing to overall better quality video assessment. 
This cross-task synergy facilitates a more robust evaluation of outputs across both image and video modalities in tasks involving understanding and generation. It inspires the development of a unified multimodal reward model that yields more precise reward signals for preference optimization.

To this end, we propose \textbf{\ourmethod}, the first unified reward model for assessing multimodal understanding and generation, capable of both pairwise ranking and pointwise scoring, which can be utilized for preference alignment on diverse vision tasks.

\input{table/recent_reward_model}

As illustrated in Fig. \ref{fig:pipeline}, our fine-tuning pipeline includes three key stages: \textbf{(1)} First, we construct a large-scale human preference dataset that spans both image and video generation/understanding tasks and develop \ourmethod based on this dataset. \textbf{(2)} Next, we employ \ourmethod to automatically construct high-quality preference pair data by selecting the outputs of specific baselines, such as Vision Language Models (VLM) and diffusion models, through multi-stage filtering, \ie pair ranking and point sifting. \textbf{(3)} Finally, we use these preference pairs to align the outputs of these models with human preferences via direct preference optimization.
Our experiments show that learning multiple visual tasks together yields significant reciprocal benefits, enhancing performance in each individual domain. By implementing our pipeline across both vision understanding and generation baselines, we observe notable improvements in each domain.

\textbf{Contributions}:
(1) We construct a large-scale human preference dataset that spans diverse vision tasks and develop \ourmethod, the first unified reward model for assessing multimodal understanding and generation.
(2) We propose a general pipeline for both vision understanding and generation model preference alignment, which remains an underexplored area in current research. Extensive experiments demonstrate its effectiveness in improving the performance of vision models in each domain.
(3) Our experiments reveal that learning to assess image and video tasks jointly leads to a synergistic improvement in performance across different visual domains.

Through this work, we aim to expand the scope of reward models, making them more adaptable, generalizable, and effective across various visual applications.

%% file: figs/shortcut1.tex
\begin{figure}[!tb]

    \centering
    \includegraphics[width=1\linewidth]{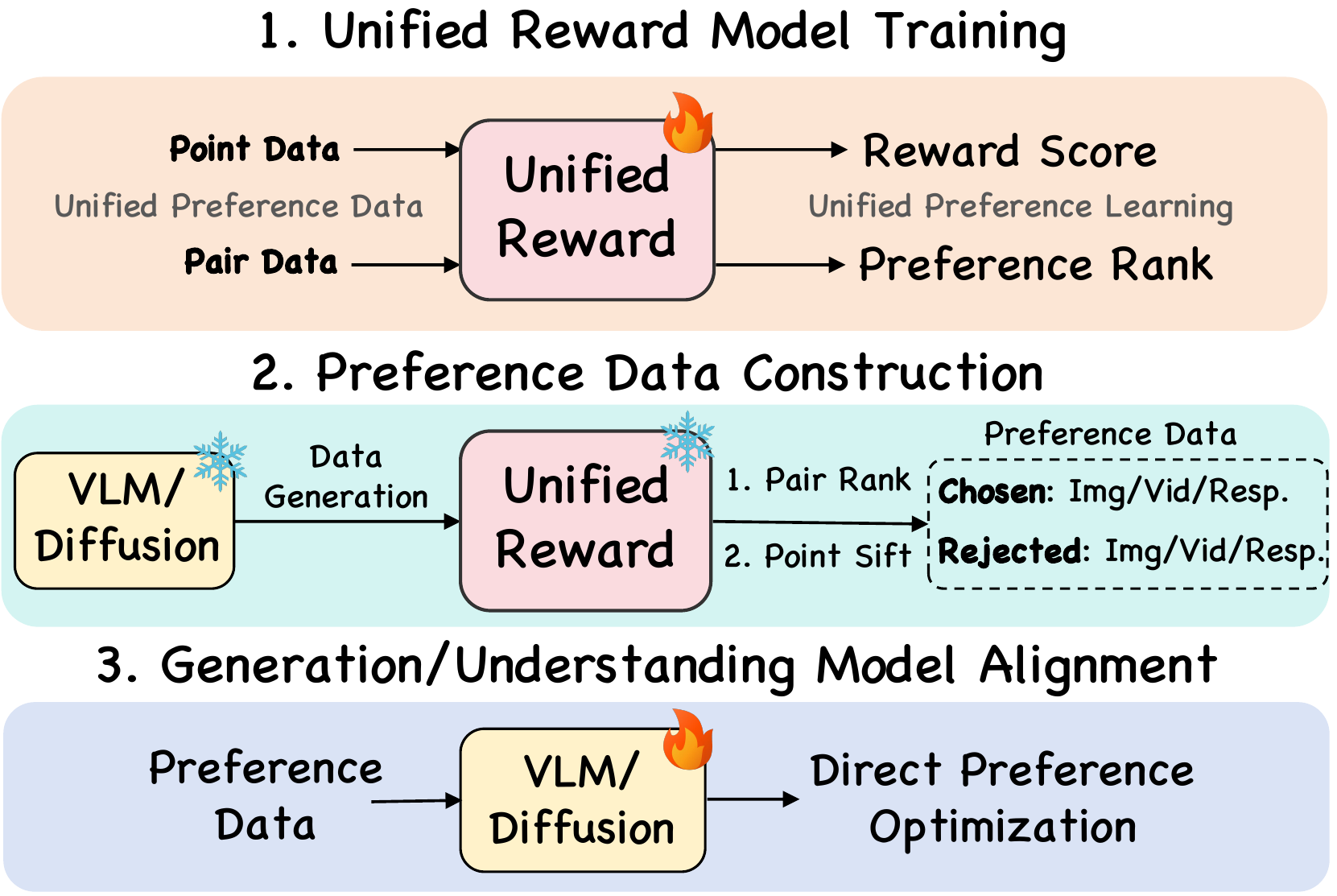}
    \caption{\textbf{Overview of UnifiedReward for Multimodal Understanding and Generation Alignment.} The pipeline includes three steps: (1) Unified Reward Model Training, (2) Preference Data Construction, and (3) Generation/Understanding Model Alignment.}

    \label{fig:pipeline}
\vspace{-0.4cm}
\end{figure}

%% file: table/recent_reward_model.tex
\begin{table}[t]
\setlength{\tabcolsep}{1.5pt}
\centering
\caption{\textbf{Comparison of Our Reward Model with Recent Approaches.} \ourmethod is capable of assessing both image and video understanding and generation. ``Pair'' and ``Point'' refer to ``Pair Ranking'' and ``Point Scoring''.}

\resizebox{\columnwidth}{!}{%
\begin{tabular}{lccccc}
\toprule
Reward Model & Method & \makecell[c]{Image\\ Generation} & \makecell[c]{Image\\ Understand} & \makecell[c]{Video\\ Generation} & \makecell[c]{Video\\ Understand} \\ \midrule
PickScore'23 \cite{pickscore}&Point&\checkmark  & &  &  \\ 
HPS'23 \cite{hps}&Point&\checkmark  & &  &  \\ 
ImageReward'23 \cite{hps}&Point&\checkmark  & &  &  \\ 
LLaVA-Critic'24 \cite{llava-critic}&Pair/Point&  & \checkmark&  &  \\
VideoScore'24 \cite{LiFT} & Point&  &  & \checkmark &  \\ 
LiFT'24 \cite{LiFT} & Point&  &  & \checkmark &  \\ 
VisionReward'24 \cite{visionreward} & Point& \checkmark &  & \checkmark &  \\ 
VideoReward'25 \cite{videoreward}& Point & &  & \checkmark &  \\ 
\midrule
\textbf{UnifiedReward} & Pair/Point& \checkmark & \checkmark & \checkmark & \checkmark \\ 	\bottomrule
\end{tabular}%
}

\label{tab:reward_model_comparison}
\vspace{-0.2cm}
\end{table}

%% file: 2.related_work.tex
\section{Related Work}
\textbf{Reward Models}
are crucial in aligning vision understanding and generation models with human preferences. Traditional methods \cite{huang2023t2i,liu2024evalcrafter,huang2024vbench}
for evaluating vision quality and semantic consistency rely on metrics such as FID \cite{heusel2017gans} and CLIP scores \cite{radford2021learning}. Despite their effectiveness, they are limited in their ability to capture human preferences. Therefore, recent studies \cite{xu2023imagereward,zhang2024learning,liang2024rich} utilize human preference data to fine-tune CLIP, enabling them to better predict and align with human evaluations. 
With the advent of VLMs \cite{achiam2023gpt,wang2024qwen2}, their robust ability to align visual and textual data makes them promising candidates for reward modeling.
These models can be adapted into two main categories based on their capabilities: understanding assessment models \cite{llava-critic,zang2025internlm}, which are designed exclusively for evaluating visual understanding tasks, and generation assessment models \cite{LiFT,visionreward,videoreward,videoscore}, which focus on assessing visual synthesis quality. 

However, these reward models are typically designed for specific tasks, as illustrated in Tab. \ref{tab:reward_model_comparison}, restricting their ability to adapt to diverse visual understanding and generative tasks. In this work, we propose the first unified reward model for both image and video understanding and generation assessment, which is more adaptable, generalizable, and effective across various visual applications.

\noindent\textbf{Preference Learning for VLM/Diffusion} is widely utilized to enhance their image and video understanding/generation performance. In video understanding, prior works have explored reinforcement learning with human feedback to refine reward models for factuality assessment \cite{sun2023aligning}, while \cite{ahn2024tuning,li2025temporal,llava-houd-dpo} have used reinforcement learning on AI-generated feedback to enhance video LMMs. 
For image understanding, researchers investigate Direct Preference Optimization (DPO) as an alternative approach to preference modeling. \cite{ahn2024tuning,gunjal2024detecting} apply DPO to refine rewards distilled from GPT-4V across different model outputs, while \cite{zhao2023beyond} constructs preference datasets by generating positive and negative sample pairs using ChatGPT, informed by detailed image descriptions. Similar methods have been applied to image generation \cite{diffusiondpo,lee2023aligning,visionreward} and video generation \cite{videodpo,videoreward,LiFT,visionreward,furuta2024improving,t2v-turbo,yuan2024self,zhang2024onlinevpo}, using reward models or human preference data to align pre-trained diffusion models. 

However, these methods rely on task-specific reward models, and no unified reward model has been developed for preference learning across both image and video generation and understanding tasks. This limits the generalizability and efficiency of reward-based alignment. Our work investigates the effectiveness of joint learning to assess multiple visual tasks, demonstrating that cross-task synergy enhances the evaluation capabilities across each domain.


%% file: 3.method.tex
\section{Method}

\subsection{Overview}
This work aims to develop a unified reward model for vision model preference alignment. Existing studies typically develop specialized reward models for specific tasks as shown in Tab. \ref{tab:reward_model_comparison}, which restricts their adaptability across diverse visual applications. Furthermore, we intuitively argue that jointly learning multiple visual tasks can create a mutually reinforcing effect, yet this remains an underexplored area. To this end, this work proposes \ourmethod, the first unified reward model for multimodal understanding and generation assessment, enabling both pair ranking and point scoring. It is then leveraged for aligning Vision-Language Models (VLMs) and Diffusion model alignment, enabling more robust and adaptable preference learning across diverse visual tasks. 
\input{figs/dataset1}

Our pipeline is illustrated in Fig. \ref{fig:model}. Specifically, we first construct a large-scale, unified preference dataset (Sec. \ref{sec:dataset_construction}) and train our \ourmethod model on this dataset (Sec. \ref{sec: reward_train}). Then, we curate preference datasets for VLMs and diffusion models by applying pair ranking and point sifting on their outputs (Sec. \ref{sec:data_construct}). These curated datasets are subsequently used for Direct Preference Optimization (DPO) (Sec. \ref{sec:alignment}), effectively aligning models with human preferences.

\subsection{Unified Reward Model Training}

\subsubsection{Unified Preference Dataset Construction} \label{sec:dataset_construction}
A comprehensive human preference dataset that spans multiple vision-related tasks is essential for training a unified reward model. However, existing human feedback datasets, such as \citep{LiFT, videodpo, llava-critic}, are typically designed for specific tasks, limiting their generalizability. Currently, there is no human preference dataset that comprehensively covers both visual understanding and generation tasks, highlighting the need for a more versatile dataset.
To bridge this gap, we integrate existing datasets and preprocess them to construct the first large-scale unified human preference dataset, which consists of approximately 236K data samples covering both image and video understanding and generation tasks. The detailed statistics and visualized distributions of the dataset are presented in Fig. \ref{fig:dataset} and Tab. \ref{tab:dataset}, respectively. We will elaborate on the data construction process for each task in the following.

\input{figs/model1}
\noindent\textbf{Image Generation.} \textit{EvalMuse} \citep{han2024evalmuse} consists of 4K prompts, each with multiple images generated by different models. Each image is evaluated by at least three annotators, who provide an overall score (1-5) and element-wise labels indicating whether specific elements are present.
For pointwise score learning, we compute the final score as the average of all ratings. An element is considered generated if at least two annotators agree; otherwise, it is marked as not generated.
We integrate the overall score and element-wise labels as assessment answers for reward model learning. For pairwise ranking, we select the images with the highest and lowest average score from the same prompt as a ranking pair. \textit{Human Preference Dataset (HPD)} \citep{HPD} contains 700K human preference votes. For each prompt, two images generated by different models are provided, each with its respective vote count. In our work, we directly use the vote counts to construct pairwise ranking data, ranking the image with more votes as the preferred one. 
\textit{Open-Image-Preferences (OIP)} $ \footnote{https://huggingface.co/datasets/data-is-better-together/open-image-preferences-v1-binarized} $ contains 7.4K text-to-image preference pairs, which are directly used in this work.
\textbf{Image Understanding.} \textit{LLava-Critic-113K} \citep{llava-critic} consists of 40K pointwise score and 73K pairwise ranking data samples for image understanding assessment learning. From this dataset, we select 25K samples for each of pairwise ranking and pointwise scoring learning.
\textbf{Video Generation.} \textit{VideoDPO} \citep{videodpo} includes 10K synthesized video pairs for text-to-video model DPO. We directly use this dataset for our pairwise ranking learning in video generation. \textit{LiFT-HRA} \citep{LiFT} and \textit{VideoFeedback} \citep{videoscore} provide extensive human feedback for pointwise scoring of synthesized videos, which we directly incorporate into our work. 
\textbf{Video Understanding.} \textit{ShareGPTVideo-DPO} \citep{sharegptvideo} contains 17K video understanding DPO data, where each response in a pair is assigned an evaluation score. We directly use the pair data for pairwise ranking learning, while the individual response scores are extracted for pointwise scoring learning.
\input{table/dataset}

For pairwise ranking datasets, we standardize the answer format as ``image/video/response X is better than image/video/response Y'', where ``X'' and ``Y'' represent the assigned indices. If the dataset includes evaluation justifications \citep{llava-critic,LiFT}, we retain them to allow the model to learn from human reasoning.
For pointwise scoring, we do not enforce a unified response format or score range, allowing the model to learn from diverse rating styles and scoring systems across different datasets. To ensure alignment between evaluation criteria and responses, we adjust instruction prompts accordingly. 

As shown in Fig. \ref{fig:dataset}, the training data for video generation pairwise ranking assessment is relatively limited compared to other tasks, but we believe that the synergistic effect of multitask learning can alleviate this deficiency. Overall, our dataset provides a diverse and comprehensive collection of human preferences, covering both pairwise ranking and pointwise scoring across image and video understanding/generation tasks. This enables effective reward model training, ensuring robust performance across multimodal understanding and generation applications.

\subsubsection{Unified Preference Learning}\label{sec: reward_train}
Based on the comprehensive datasets, we fine-tune a pre-trained VLM \citep{llava-ov} with strong vision understanding capabilities to develop \ourmethod, jointly training it across diverse vision tasks. Instead of learning evaluation from scratch, we integrate assessment ability as an additional discriminative skill, leveraging the model’s existing visual comprehension to enhance its evaluation performance across various tasks.

Fig. \ref{fig:model} (top) illustrates our training process. Specifically, for multimodal generation evaluation, our model takes vision tokens, instruction input, and a caption as input. In contrast, for multimodal understanding, the caption is replaced by a question and the corresponding response(s), aligning the input format with the respective task requirements.
The model is trained to predict the pointwise score or pairwise ranking based on the criteria specified in the instruction prompt. If the training data includes justifications, the model is also trained to generate detailed explanations to support its evaluations. During training, the optimization objective is standard cross-entropy loss, but it is computed only on the model's predicted answer. 

After training our \ourmethod, we leverage it for preference alignment in multimodal understanding and generation models. This process consists of two sequential steps, \ie Preference Data Construction and Generation/Understanding Model Alignment.
The following sections provide a detailed explanation of each step.

\subsection{Preference Data Construction} \label{sec:data_construct}
The quality of preference alignment data directly determines the effectiveness of model alignment. Existing methods \citep{LiFT,videoreward,llava-critic} are often limited to a single evaluation strategy, either assigning pairwise rankings or pointwise scores to model outputs for preference data construction. In contrast, this work leverages both pairwise ranking and pointwise scoring capabilities of \ourmethod, enabling a higher quality preference data construction pipeline, as illustrated in Fig. \ref{fig:model} (bottom left). 

Specifically, our pipeline includes three sequential steps: (1) \textbf{Data Generation}. Given an image/video-question pair (or generation prompt), a VLM (or diffusion model) generates multiple candidate outputs 
$\{ O_1, O_2, \dots, O_N \}$. These outputs serve as the initial pool for followed preference data filtering. (2) 
\textbf{Pair Ranking}.
Given \textit{N} outputs, we group them into \textit{N}/2 pairs and use our model to perform pairwise ranking for each pair. Then, we classify these ranked pairs into a chosen list \(\mathcal{C} = \{ O^c_1, O^c_2, \dots, O^c_{N/2} \} \) and a rejected list \(\mathcal{R} = \{ O^r_1, O^r_2, \dots, O^r_{N/2} \}\). 
(3) \textbf{Point Sifting}. 
Finally, we apply our model to assign pointwise scores to all outputs in both the chosen list and the rejected list. The final preference data pair is determined as:
\[
(O^*_c = \arg\max_{O \in \mathcal{C}} S(O), \quad O^*_r = \arg\min_{O \in \mathcal{R}} S(O)),
\]
where \( S(O) \) represents the pointwise score assigned by our model, \( O^*_c \) is the most preferred output and \( O^*_r \) is the least preferred output.

By combining pairwise ranking and pointwise scoring, the final preference data could provide a high-quality and reliable preference signal, effectively capturing both relative comparisons and absolute quality assessments. 


\subsection{Generation/Understanding Model Alignment} \label{sec:alignment}
After constructing the preference data, we leverage it for multimodal generation and understanding model alignment using DPO, which enables models to align their outputs with human preferences without explicit reward modeling, optimizing directly based on ranked preference pairs.

\textbf{DPO for Multimodal Generation}.
For multimodal generation tasks, diffusion \citep{ho2020denoising} is widely used due to their strong capability in generating high-quality and diverse outputs across image and video synthesis. Therefore, we apply DPO on diffusion models to align their outputs with human preferences. 

Given the constructed preference pair dataset \(\mathcal{D}_{Gen} = \{ (x^w_0, x^l_0)_{i} \}_{i=1}^{M} \), where \(x^w_0\) and \(x^l_i\) represents the preferred generated sample and the less preferred sample respectively, \textit{M} denotes the number of samples, we optimize the diffusion model by comparing the noise prediction differences between the fine-tuned model and pre-trained reference model \citep{diffusiondpo}:
\begin{equation}
\begin{aligned}
L(\theta) = -\mathbb{E}_{(x^w_0, x^l_0) \sim \mathcal{D}_{Gen}, \, t \sim \mathcal{U}(0,T), \, x^w_t \sim q(x^w_t | x^w_0), \, x^l_t \sim q(x^l_t | x^l_0)} \\
\log \sigma \Bigg( -\beta_g T \omega(\lambda_t) \Big( 
\|\epsilon^w - \epsilon_{\theta}(x^w_t, t)\|_2^2 - \|\epsilon^w - \epsilon_{\text{ref}}(x^w_t, t)\|_2^2  \\
- \Big( \|\epsilon^l - \epsilon_{\theta}(x^l_t, t)\|_2^2 - \|\epsilon^l - \epsilon_{\text{ref}}(x^l_t, t)\|_2^2 \Big) 
\Big) 
\Bigg), \nonumber
\end{aligned}
\end{equation}
where \( x^w_t \) and \( x^l_t \) are the noisy latents derived from \( x^w_0 \) and \( x^l_0 \) at timestep \( t \), respectively.  
\( \epsilon_{\theta}(x_t^*, t) \) and \( \epsilon_{\text{ref}}(x_t^*, t) \) denote the predicted noise from the fine-tuned and pre-trained reference diffusion models, respectively.  
\( \beta_g \) is a temperature hyperparameter controlling optimization strength,  $\sigma$ is the logistic function,
\( \lambda_t \) represents the signal-to-noise ratio,  
and \( T \omega(\lambda_t) \) is a weighting function, which is treated as a constant equal to \( \beta_g \) in this work.

This loss encourages the fine-tuned diffusion model to reduce the denoising error for preferred samples while increasing it for less preferred ones, thereby improving the generation quality.

\textbf{DPO for Multimodal Understanding}.
Similar to generation alignment, we apply DPO to adjust the model’s response preference for multimodal understanding models, \ie VLMs. Given an input \( x \) (\eg an image/video-question pair) with a preferred response \( y_w \) and a less preferred response \( y_l \) from preference pair dataset $\mathcal{D}_{\text{Und}}$, the optimization is formulated as:
\begin{equation}
\begin{aligned}
\mathcal{L}(\theta) = 
- \mathbb{E}_{(x, y_w, y_l) \sim \mathcal{D}_{\text{Und}}} 
\Bigg[ \beta_{u} \log \sigma \Bigg(  \log 
\frac{\pi_{\theta}(y_w | x)}{\pi_{\text{ref}}(y_w | x)} \\
-  \log \frac{\pi_{\theta}(y_l | x)}{\pi_{\text{ref}}(y_l | x )} 
\Bigg) \Bigg] \, ,
\end{aligned} \nonumber
\end{equation}
where \( \pi_{\theta}(y_* | x )\) and \( \pi_{\text{ref}}(y_* | x )\) are the response probability under the fine-tuned model and reference model, respectively. \( \beta_u \) is a hyperparameter that controls optimization sensitivity.

This loss encourages the VLM to increase the likelihood of generating preferred responses while decreasing it for less preferred ones, thereby improving the model’s alignment with human preferences and enhancing reasoning quality.


%% file: figs/dataset1.tex
\begin{figure}[!tb]

    \centering
    \includegraphics[width=1\linewidth]{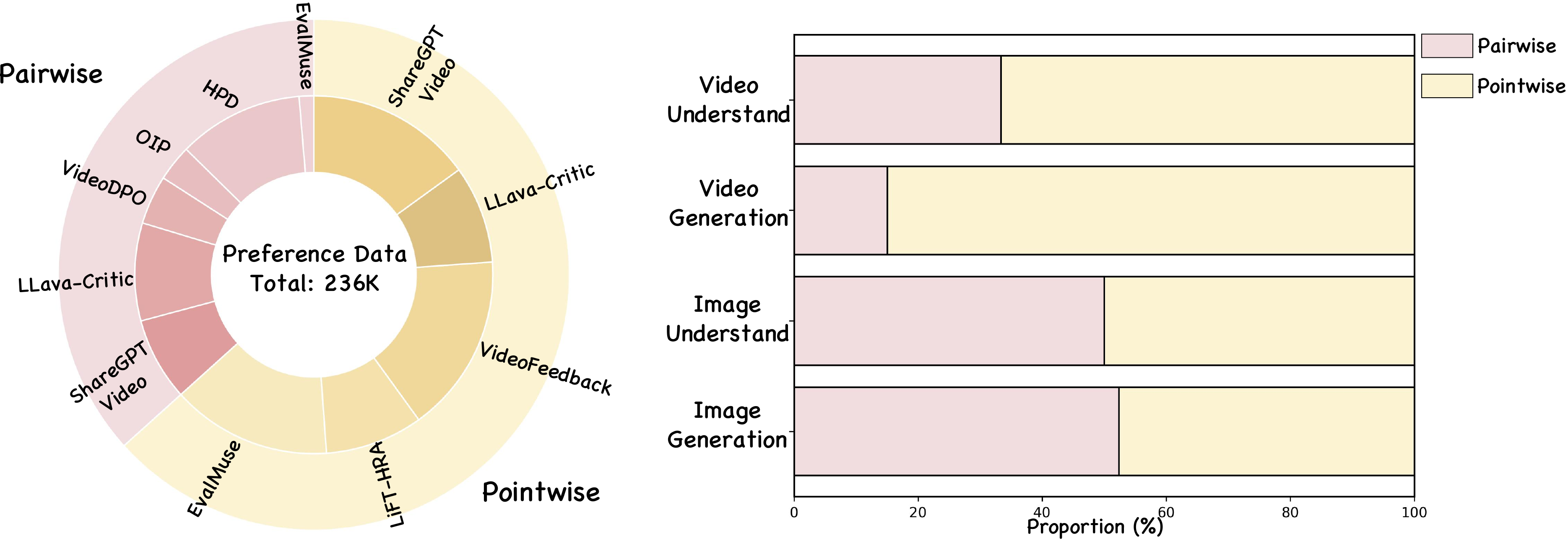}
    \caption{\textbf{Visualization of Statistical Results}. This figure presents the distribution of our constructed unified preference dataset, along with the pairwise and pointwise distributions for each task.}

    \label{fig:dataset}
\vspace{-0.3cm}

\end{figure}

%% file: figs/model1.tex
\begin{figure*}[!th]

    \centering
    \includegraphics[width=1\linewidth]{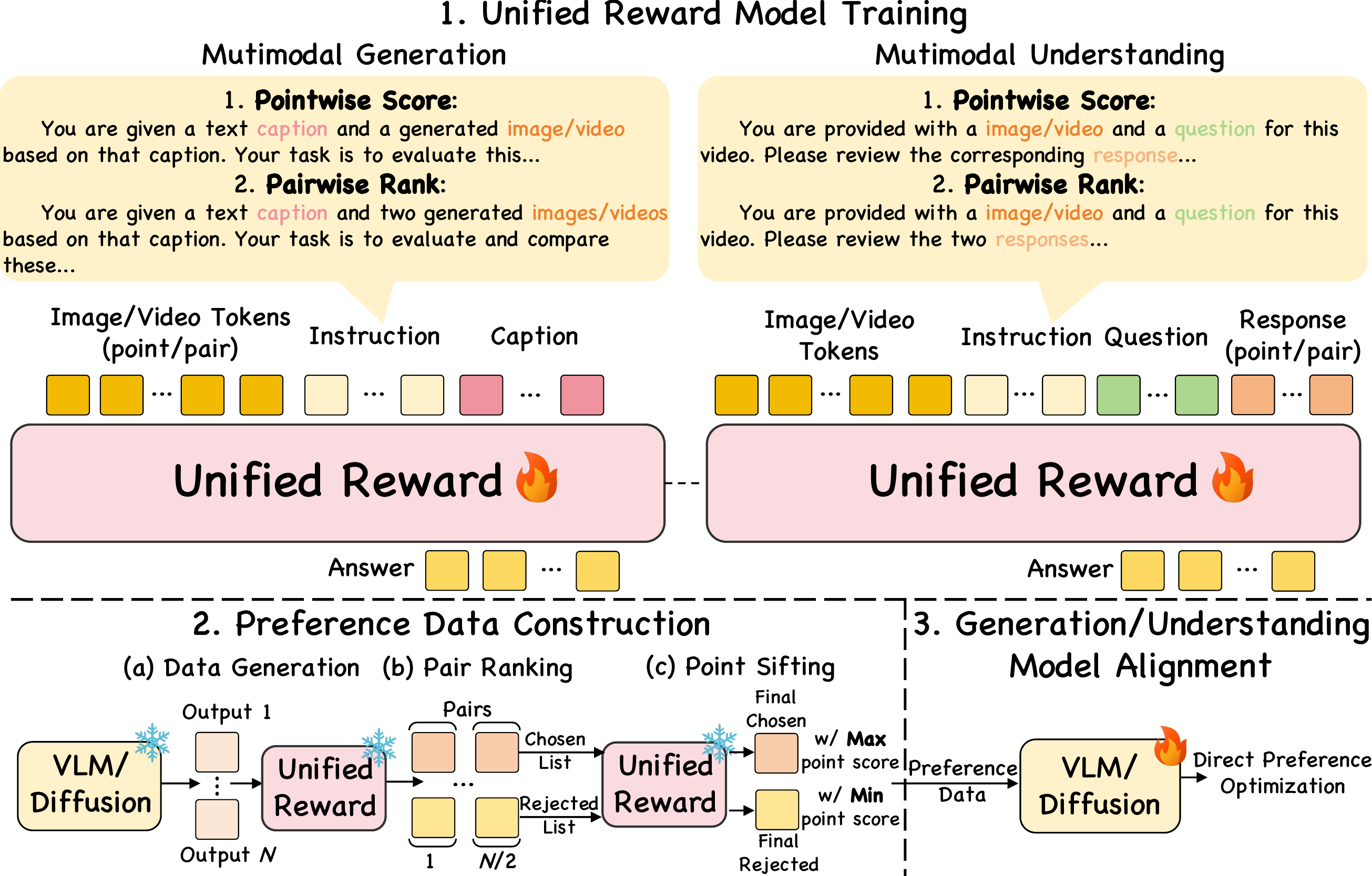}
    \caption{\textbf{Method Overview.} 
(1) \textit{Unified Reward Model Training}: train a unified reward model for both multimodal generation and understanding assessment using pointwise scoring and pairwise ranking strategy.  
(2) \textit{Preference Data Construction}: use the trained reward model to construct high-quality preference data through three steps: (a) data generation from vision models, (b) pairwise ranking to divide the chosen and rejected outputs, and (c) pointwise filtering to refine the chosen and rejected samples.  
(3) \textit{Generation/Understanding Model Alignment}: the constructed preference data is then used to align vision models with human preference via Direct Preference Optimization (DPO).}
    \label{fig:model}
\vspace{-0.4cm}
    
\end{figure*}

%% file: table/dataset.tex
\begin{table}[!t]
\setlength{\tabcolsep}{2.5pt}
\centering
\caption{\textbf{Training Datasets for Image and Video Generation/Understanding Assessment.} ``*'' indicates the dataset is preprocessed in our work.}

\resizebox{\columnwidth}{!}{%
\begin{tabular}{l|lll|c}
\toprule
 & Task & Method & Dataset & Size \\
\midrule
\multirow{6}{*}{Image} & \multirow{3}{*}{Generation} & \multirow{3}{*}{Pair} & EvalMuse* & 3K \\
& &  & HPD* & 25.6K \\
& &  & OIP & 7.4K \\
& & Point & EvalMuse* & 32.7K \\
\cline{2-5}
& \multirow{2}{*}{Understanding} & Pair & LLaVA-Critic & 25K \\
& & Point & LLaVA-Critic & 25K \\
\midrule
\multirow{5}{*}{Video} & \multirow{3}{*}{Generation} & Pair & VideoDPO & 10K \\
& &\multirow{2}{*}{Point} & LiFT-HRA & 20K \\
& & & VideoFeedback & {36.6K} \\
\cline{2-5}
 & \multirow{2}{*}{Understanding} & Pair & ShareGPTVideo & 17K \\
& & Point & ShareGPTVideo* & {34K} \\
\bottomrule
\end{tabular}%
}
\vspace{-0.4cm}

\label{tab:dataset}
\end{table}

%% file: 4.exp.tex
\section{Experiments}
\input{table/vlrewardbench}

\subsection{Implementation Details}
\textbf{Reward Model}. We adopt the pre-trained \textit{LLaVA-OneVision} 7B (OV-7B) \citep{llava-ov} as the base architecture for \ourmethod. Training is conducted on 8 H100 GPUs with a batch size of 2, gradient accumulation steps of 16, a learning rate of \(2.5 \times 10^{-6}\), and a warm-up ratio of \(0.3\). Inference remains efficient: $ \sim $1s for direct answers and $ \sim $3s with brief rationales, with no additional overhead compared to the base model. We additionally train \ourmethod on \textit{Qwen2.5-VL} \citep{bai2025qwen2} to verify the robustness of \ourmethod across different baselines. 

\textbf{Multimodal Understanding DPO}. Based on \ourmethod, we apply DPO to \textit{LLaVA-OneVision} 7B \citep{llava-ov} and \textit{LLaVA-Video} \citep{llava-video} to enhance their performance in image and video understanding, respectively. We use a batch size of 1, gradient accumulation steps of 16, a learning rate of \(5 \times 10^{-7}\), and set \( \beta_u = 0.1 \).

\textbf{Multimodal Generation DPO}. For image and video generation DPO, we use \textit{SDXL-Turbo} \citep{sdxl} and \textit{T2V-Turbo} \citep{t2v-turbo}, respectively. The parameter \( \beta_g \) is set to 5000, with batch sizes of 32 for \textit{SDXL-Turbo} and 16 for \textit{T2V-Turbo}. 
We construct 10K preference data for video generation DPO and 14k for other DPO tasks. The number of candidate outputs \textit{N} is set to 10. All models are trained for 3 epochs. 
\input{figs/qualitative_video1}

\textbf{Evaluations}.
\noindent\textbf{Multimodal Understanding}: We evaluate the image and video understanding assessment of \ourmethod on VLRewardBench \cite{vlrewardbench} and ShareGPTVideo \cite{sharegptvideo} (1K samples for testing), respectively. 
\noindent\textbf{Multimodal Generation}: GenAI-Bench \cite{genai-bench} includes both image and video generation reward benchmarks, which are utilized. Besides, we also employ VideoGen-RewardBench \cite{videoreward} as the video generation assessment benchmark. 
\noindent\textbf{DPO}: For image understanding, LLaVABench \citep{llava-bench}, WildVision \citep{lu2024wildvision}, LLaVABench-Wilder \citep{llava-bench-wilder}, LiveBench \citep{white2024livebench}, MMHal \citep{mmhal}, MMBench \citep{liu2024mmbench}, MME \citep{zhang2021mme}, MathVista \citep{lu2023mathvista}, DocVQA \citep{mathew2020document}, and TextVQA \citep{singh2019towards} are employed. We use LMMs-Eval \citep{lmms_eval2024} toolkit to evaluate.
For video understanding, we employ ``\textit{gpt-3.5-turbo-1106}'' for MSRVTT \cite{msrvtt}, MSVD \cite{hendria2023msvd}, TGIF \cite{li2016tgif} evaluation, while using the VLMEvalKit \citep{duan2024vlmevalkit} toolkit for evaluate LongVideoBench \cite{wu2025longvideobench}, MLVU \cite{zhou2024mlvu} and VideoMME \cite{fu2024video} evaluation. 
For image generation evaluation, we generate images conditioned on captions from
the Partiprompt \cite{yu2022scaling} and HPSv2 \cite{hps} benchmarks (1632
and 3200 captions respectively) and utilize the image reward models, \ie PickScore \cite{pickscore}, HPSv2 \cite{hps} and ImageReward \cite{xu2023imagereward} for quality assessment. VBench \cite{huang2024vbench} is used for video generation assessment.

\input{figs/qualitative_image1}

\subsection{Reward Model Comparison Results}
\textbf{Image Understanding}. We compare our method with LLaVA-Critic \citep{llava-critic}, as well as two closed-source models \citep{team2024gemini,islam2024gpt}.
The experimental results, shown in Tab. \ref{tab:vlreward}, indicate that our method outperforms baselines in most metrics, \eg macro accuracy, which demonstrates the superiority of our method in image understanding assessment. 
For \textbf{Video Understanding}, we explore the effectiveness of multi-task learning in video understanding assessment on ShareGPTVideo, which will be analyzed in Sec. \ref{sec:multi_task_learning}. In \textbf{Image Generation} assessment, we compare our method with both traditional and state-of-the-art approaches \citep{pickscore,hps,xu2023imagereward,visionreward}.
The results are presented in Tab. \ref{tab:benchmark_comparison}.
Notably, the latest work, VisionReward, supports reward modeling for both image and video generation. However, it trains separate models for each task using their respective datasets, whereas our approach jointly learns multiple tasks within a unified framework, leading to relatively better performance.
For \textbf{Video Generation}, we compare our method with the latest approaches \citep{videoscore,LiFT,visionreward,videodpo}.
As shown in Fig. \ref{fig:dataset}, our training data for video generation assessment is relatively limited. However, as demonstrated in Tab. \ref{tab:benchmark_comparison}, our method excels across all metrics when compared to all baselines, highlighting that multitask learning not only mitigates the issue of insufficient training data but also enhances the learning effectiveness for video generation assessment. 


\input{table/image_video_generation}

\subsection{DPO Comparison Results}

\input{table/video_understanding_dpo}

\noindent\textbf{Image Understanding}. 
We compare our method with LLaVA-Critic by employing the same image-question pair source \citep{2023llavarlhf} to construct preference data for OV-7B, ensuring a fair comparison. The results, presented in Tab. \ref{tab:image_und_dpo}, 
demonstrate that DPO using our method consistently outperforms the baseline across all benchmarks. For instance, our method achieves a 3.4\% improvement on LLaVABench, highlighting its superior effectiveness.
\input{table/image_understanding_dpo}
\noindent\textbf{Video Understanding}. We extract prompts from ShareGPTVideo-DPO \citep{sharegptvideo} to construct preference data for LLaVA-Video-7B \citep{llava-video}, sharing the same video-question pair source as LLaVA-Houd-DPO \citep{llava-houd-dpo}. To evaluate the effectiveness, we compare our \ourmethod-based DPO with Houd-DPO and the latest TPO \citep{li2025temporal}. The results, presented in Tab. \ref{tab:vid_und_dpo}, demonstrate the superiority of our approach. Notably, our method significantly outperforms the baselines on MSRVTT, MSVD, and TGIF, demonstrating its effectiveness in video understanding. For the other three multi-choice question datasets, although our DPO data do not include this question type, this does not lead to any negative impact. Our performance still remains comparable to the baselines, indicating the robustness and generalization ability of our approach.
For \textbf{Image Generation}, we extract prompts from Pick-a-Pic \citep{pickscore}, to construct preference data. As shown in Tab. \ref{tab:gen_dpo_comparison} (A), training on the constructed data using our \ourmethod achieves better performance compared to directly training on the original dataset. This demonstrates the effectiveness of our approach in refining preference data for improved model alignment. The qualitative comparison results are shown in Fig. \ref{fig:qualitative_image}.

\input{table/image_generation}

For \textbf{Video Generation}, we compare our method with VideoDPO \citep{videodpo}, using the same prompt source for preference data construction. The results in Tab. \ref{tab:gen_dpo_comparison} (B) demonstrate our superiority in enhancing both generation quality and semantic consistency, highlighting the effectiveness of our approach. 
The qualitative comparison results are shown in Fig. \ref{fig:qualitative_video}.


\subsection{Discussion} \label{sec:multi_task_learning}
\subsubsection{\textbf{Multi-task Assessment Learning}}
This work intuitively argues that visual tasks are inherently interconnected, and jointly learning multiple visual tasks may create a mutually reinforcing effect. Therefore, we explore the effectiveness of multi-task learning on the reward model.
Specifically, for each task, we employ different training data configurations to train the model, investigating the impact of jointly learning across different modalities (image and video) and tasks (understanding and generation). For example, for the image understanding task, we design three training configurations to investigate the impact of multi-task learning: (1) training solely on image understanding assessment, (2) jointly learning image understanding and image generation assessment, and (3) jointly learning image understanding and video understanding assessment. The results are presented in Tab. \ref{tab:vlreward}.
Notably, our findings indicate that multi-task learning significantly enhances the model’s overall performance compared to training on a single task. For instance, jointly training on both image and video understanding tasks improves overall accuracy and macro accuracy by 5.3\% and 8.3\%, respectively, compared to training solely on image understanding. 
Results for other tasks are presented in Tabs. \ref{tab:sharegpt} and \ref{tab:benchmark_comparison}, which consistently demonstrate its effectiveness.
These results highlight the benefits of leveraging shared knowledge across different visual tasks, leading to a more robust and generalizable reward model.
\input{table/sharegptvideo}

\subsubsection{\textbf{Cross-Task Synergy Beyond Scaling Data}}
To verify that the gains of \ourmethod do not simply stem from a larger training set, we introduce a \textbf{budget-matched control} in Tab.~\ref{tab:synergy_transfer} (A). Specifically, \textit{Single-task (native)} denotes single-domain reward models trained on their original task-specific data, while \textit{Single-task (step-matched)} denotes the same models further oversampled to match the total number of update steps used by \ourmethod. Despite this budget matching, our model still achieves the best performance across all evaluation axes. This shows that its advantage is not merely due to having more training data or longer optimization, but arises from positive cross-task synergy between understanding and generation. We further analyze \textbf{directional transfer} by training the reward model on a single domain and evaluating across all domains (Tab. \ref{tab:synergy_transfer} (B)). The results show clear cross-task promotion: understanding-centric training also improves generation evaluation, indicating that transfer is not limited to the source domain. At the same time, single-domain training generalizes best within its own modality, while still providing consistent positive gains to other modalities. In contrast, \ourmethod remains consistently strong across all targets, suggesting that unified multi-task learning strengthens positive transfer while reducing modality-specific overfitting.
\input{table/synergy_transfer}
\input{figs/flux_grpo_compare1}
\input{figs/sdxl_dpo_compare1}

\subsubsection{\textbf{Robustness on Different Baselines}}
To further demonstrate robustness across base models, we additionally train \ourmethod on Qwen2.5-VL \citep{bai2025qwen2}. As shown in Tab. \ref{tab:llava_qwen}, the same improvement trend holds on this stronger backbone. We also observe that larger backbones deliver better overall results while preserving the advantage of \ourmethod, suggesting that our approach is compatible with stronger priors and scales reliably with model capacity.
\input{table/llava_qwen}

\subsubsection{\textbf{Impact of Imbalanced Training Data}}
We investigate the impact of cross-task data imbalance in Tab. \ref{tab:imbalance_data}. Starting from a balanced allocation, increasing the amount of non-video-generation data while keeping the video-generation data fixed improves performance on the over-represented tasks, but consistently degrades video-generation performance. This suggests that underrepresented tasks are more susceptible to being overwhelmed during joint optimization. Once the video-generation data is rebalanced, its performance recovers, while the other tasks remain largely stable. These results highlight the importance of balanced sampling for maintaining mutual gains across tasks.
\input{table/imbalance_data}

\subsubsection{\textbf{Preference Signal Source Comparison}}
We compare different preference signal sources for constructing SDXL \citep{sdxl} DPO pairs. To ensure a fair comparison, we keep the same DPO backbone and training pipeline, and only replace the signal source. As shown in Tab. \ref{tab:data_construction_ablation} (A), \ourmethod consistently outperforms both GPT-4o and the image-generation-only reward model (under matched training budgets) across evaluation dimensions, indicating that our reward signal provides stronger supervision than closed-source and single-domain alternatives. The qualitative cases in Fig. \ref{fig:sdxl_dpo_compare} show the same trend, with better prompt alignment, cleaner compositions, and fewer visible artifacts.

\subsubsection{\textbf{Data Construction Strategy Ablation}}
We further conduct an ablation on our proposed two-stage preference data construction strategy. Tab. \ref{tab:data_construction_ablation} (B) shows that the full two-stage pipeline consistently outperforms random selection, point-score-only, and pair-rank-only variants. This comparison indicates that the two stages play complementary roles: pair ranking provides reliable relative ordering between candidates, while point scoring filters out low-quality responses that may still survive pairwise comparison. Combining them yields cleaner and more stable preference pairs, which translates into stronger downstream alignment performance.

\input{table/data_construction_ablation}

\subsubsection{\textbf{Applied to Image Generation GRPO}}
We further test whether \ourmethod can generalize beyond DPO-style optimization pipelines by applying it to group relative policy optimization (GRPO) \citep{grpo} on FLUX.1-dev \citep{flux}. As shown in Tab. \ref{tab:grpo_flux}, reward optimization with \ourmethod consistently improves all reported evaluation views over the vanilla FLUX baseline. Compared with alternative reward signals, our method remains strongest overall, indicating that the learned reward provides stable and transferable guidance under policy optimization.
The qualitative comparison in Fig. \ref{fig:flux_grpo_compare} further supports this trend: relative to the FLUX baseline and other reward variants, samples optimized with \ourmethod show better prompt faithfulness, cleaner compositions, and more coherent fine-grained details. This consistency between quantitative and qualitative results suggests that our learned reward can serve as an effective optimization signal in various settings.
\input{table/grpo_flux}

\section{Ethical Statement}
In this work, we affirm our commitment to ethical research practices and responsible innovation. To the best of our knowledge, this study does not involve any data, methodologies, or applications that raise ethical concerns. All experiments and analyses were conducted in compliance with established ethical guidelines, ensuring the integrity and transparency of our research process.

%% file: table/vlrewardbench.tex
\begin{table}[tb]
\setlength{\tabcolsep}{6pt}
\centering
\caption{\textbf{Image Understanding Assessment.} We evaluate various aspects on VLRewardBench.}

\resizebox{\columnwidth}{!}{%
\begin{tabular}{lccccc}
\toprule
{Models} & General & Hallu. & Reason. & \makecell[c]{{Overall}\\ {Accuracy}} & \makecell[c]{{Macro}\\ {Accuracy}}\\ 
\midrule

Gemini-1.5-Pro  & 50.8 & \underline{72.5} & \underline{64.2} & \textbf{67.2} & \underline{62.5} \\ 
GPT-4o           & 49.1 & 67.6 & \textbf{70.5} & 65.8 & 62.4 \\ 

LLaVA-Critic         & 47.4 & 38.5 & 53.8 & 46.9 & 46.6 \\
\midrule
OV-7B  & 32.2 & 20.1 & 57.1 & 29.6 & 36.5 \\ 
w/ Img. Und.  & 47.6 & 38.3 & 54.5  &47.4 & 46.8 \\ 
w/ Img. Und.+Gen.  & 49.8 & 52.6 & 58.1 & 50.4 & 53.5 \\
w/ Img.+Vid. Und  & \underline{52.4} & 55.6 &  57.2 & 52.7 & 55.1 \\ 

\midrule
\textbf{UnifiedReward}     & \textbf{60.6} & \textbf{78.4} & 60.5 & \underline{66.1} & \textbf{66.5} \\ 
\bottomrule
\end{tabular}%
}
\label{tab:vlreward}
\end{table}

%% file: figs/qualitative_video1.tex
\begin{figure*}[!tb]

    \centering
    \includegraphics[width=1\linewidth]{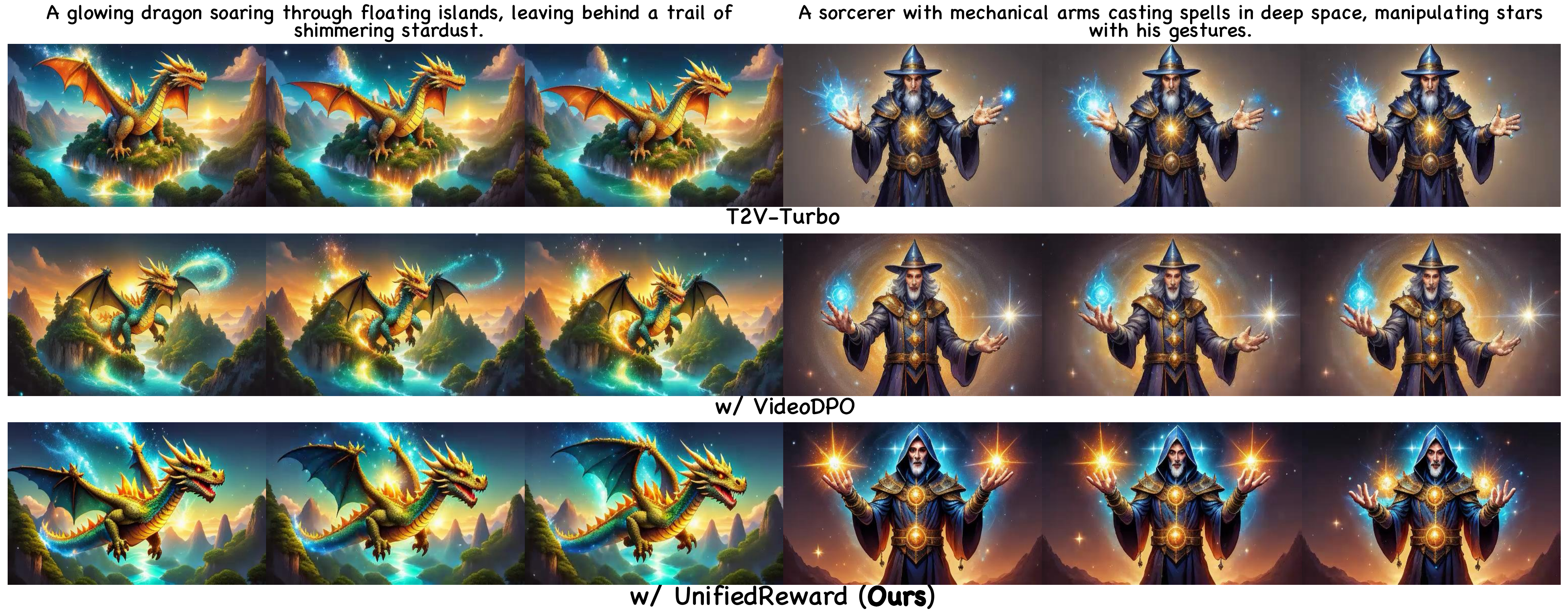}
    \caption{\textbf{Qualitative DPO Comparison on T2V-Turbo}. We compare the qualitative performance of the original T2V-Turbo, DPO trained with VideoDPO, and DPO trained with UnifiedReward.}
    \label{fig:qualitative_video}

\end{figure*}

%% file: figs/qualitative_image1.tex
\begin{figure}[!tb]

    \centering
    \includegraphics[width=1\linewidth]{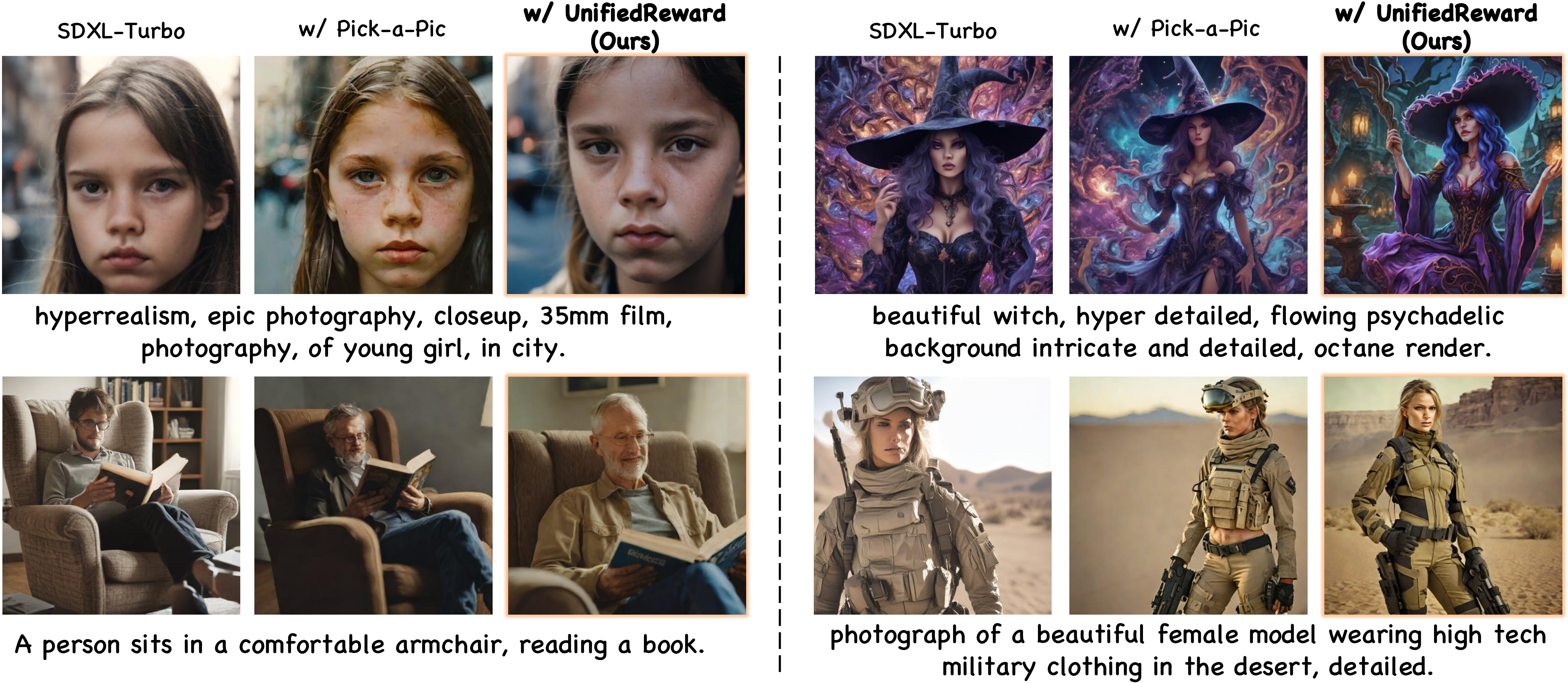}
    \caption{\textbf{Qualitative DPO Comparison on SDXL-Turbo}. We compare the qualitative performance of the original SDXL-Turbo, DPO trained on Pick-a-Pic dataset, and DPO trained with UnifiedReward.}
    \label{fig:qualitative_image}
\end{figure}

%% file: table/image_video_generation.tex
\begin{table}[tb]
\centering
\renewcommand{\arraystretch}{1.1}
\caption{\textbf{Image and Video Generation Comparison.}  ``tau'' indicates that accuracy is calculated with ties, and ``diff'' excludes tied pairs when calculating accuracy.}
\setlength{\tabcolsep}{5pt} 
\resizebox{\columnwidth}{!}{%
\begin{tabular}{lcc|ccccc}
\toprule
\multirow{3}{*}{{Method}} & \multicolumn{2}{c|}{\textbf{Image Generation}} &\multirow{3}{*}{{Method}} & \multicolumn{4}{c}{\textbf{Video Generation}} \\

 \cmidrule(lr){5-8}
& \multicolumn{2}{c|}{{GenAI-Bench}} & & \multicolumn{2}{c}{{GenAI-Bench}} & \multicolumn{2}{c}{{VideoGen-Reward}}\\
& tau& diff &  & tau & diff & tau& diff\\ 
\midrule
PickScore & \underline{53.2} & 67.2 &VideoScore & 46.2 & 70.6 & 42.1 & 49.9 \\ 
HPSv2  & 51.6 & \underline{68.4} & LiFT & 41.2  & 60.1 & 40.6 & 58.3\\ 
ImageReward  & 47.8 & 65.0 & VisionReward& \underline{52.1} & 73.1 & 57.4 &  68.2\\ 
VisionReward  & 46.8 & 66.4& VideoReward & 50.2 & 73.3 & \underline{60.1} & \underline{73.9}\\ 
\midrule
OV-7B & 39.7 & 53.2 & OV-7B & 40.8 & 51.4& 40.4 & 50.2\\ 
w/ Img. Gen. & 39.4 & 64.0& w/ Vid. Gen. & 48.2 & 69.4 & 44.3 &62.4\\ 
w/ Img. Gen.+Und. & 47.7 &65.9 & w/ Vid. Gen.+Und.  & 49.1 & 71.6& 45.1 & 64.9\\ 
w/ Img.+Vid. Gen. & 50.5 & 67.6 & w/ Img.+Vid. Gen. & 52.0 & \underline{73.6} & 53.6 &70.7\\ 

\midrule
\textbf{UnifiedReward} & \textbf{54.8} & \textbf{70.9} &\textbf{UnifiedReward}& \textbf{60.7} & \textbf{77.2} & \textbf{66.6} & \textbf{79.3}\\ 
\bottomrule
\end{tabular}%
}
\label{tab:benchmark_comparison}
\end{table}

%% file: table/video_understanding_dpo.tex
\begin{table*}[t]
\centering
\scriptsize
\renewcommand{\arraystretch}{1.1}
\caption{\textbf{Video Understanding DPO Comparison.} All methods are trained with the same settings.}
\setlength{\tabcolsep}{3.5pt} 
\resizebox{\textwidth}{!}{%
\begin{tabular}{lcccccccccccc}
\toprule
\multirow{3}{*}{{Method}} & \multicolumn{2}{c}{{MSRVTT}} & \multicolumn{2}{c}{{MSVD}} & \multicolumn{2}{c}{{TGIF}} & {LongVideoBench} & {MLVU} & \multicolumn{4}{c}{{Video-MME}}\\

\cmidrule(lr){2-3} \cmidrule(lr){4-5} \cmidrule(lr){6-7} \cmidrule(lr){10-13}
& Acc. & Score & Acc. & Score & Acc.& Score & Acc. & M-Avg. & Short & Medium & Long & Avg.\\ 
\midrule
LLaVA-Video-7B'24 & 52.8 & 3.24  & 69.7 & 3.90 & 51.9 & 3.37 &58.1&70.9&76.1&\underline{61.6}&\underline{52.3}& 63.3\\ 
w/ Houd-DPO'24 & \underline{56.8} &\underline{3.34} &\underline{72.8} & \underline{3.97} & \underline{54.9} & \underline{3.45} &58.0 &71.8 &\underline{76.3} &61.3&51.2&63.0\\ 
w/ TPO'25 & 55.0 &3.25 & 72.6 & 3.93 & 53.7 & 3.40&\underline{58.2}&\textbf{72.6}&\textbf{76.9}&\textbf{62.1}&52.1&\textbf{63.7}\\ 
\midrule
w/ \textbf{UnifiedReward} & \textbf{65.0} & \textbf{3.45} &\textbf{78.3} &\textbf{4.01} &\textbf{59.7} &\textbf{3.51} & \textbf{58.4}&\underline{72.3}&76.2&61.3&\textbf{52.5}&\underline{63.5}\\ 
\bottomrule
\end{tabular}%
}
\label{tab:vid_und_dpo}
\end{table*}

%% file: table/image_understanding_dpo.tex
\begin{table*}[t]
\setlength{\tabcolsep}{6.5pt}
\centering
\caption{\textbf{Image Understanding DPO Comparison.} We compare our method with LLaVA-Critic for DPO based on LLaVA-OneVision-7B.}

\resizebox{\textwidth}{!}{%
\begin{tabular}{lcccccccccc}
\toprule
\textbf{} & LLaVABen. & WildVision & LLaVABenWilder &LiveBen. & MMHal&MMBen&MME&MathVista&DocVQA&TextVQA\\ 
\midrule

OV-7B         & 90.3 & 54.9 & 67.8&77.1&3.19&\underline{80.9}&1994.1&62.6&\underline{87.2}&\textbf{80.1} \\ 
w/ LLaVA-Critic           & \underline{100.3} & \underline{67.3} & \underline{71.6} & \underline{84.5} & \underline{3.91} &80.5&\underline{1998.9}&\textbf{63.2}&86.98&79.2\\ 
\midrule

\textbf{w/ UnifiedReward}            & \textbf{101.4} & \textbf{67.8} & \textbf{75.0} & \textbf{85.4} & \textbf{4.01}&\textbf{81.2}&\textbf{2008.5}&\underline{62.9}&\textbf{87.4}&\underline{79.5}\\ 
\bottomrule
\end{tabular}%
}
\vspace{-0.2cm}
\label{tab:image_und_dpo}
\end{table*}

%% file: table/image_generation.tex
\begin{table}[tb]
\centering
\setlength{\tabcolsep}{4pt}
\caption{\textbf{Generation DPO Comparison.} (A) Image generation DPO evaluated by image reward metrics. (B) Video generation DPO evaluated on VBench.}

\resizebox{\columnwidth}{!}{%
\begin{tabular}{lccc}
\toprule
\multicolumn{4}{l}{\textbf{(A) Image Generation DPO (SDXL-Turbo)}} \\
{Method} & {PickScore} & {HPSv2} & {ImageReward} \\
\midrule
Baseline & 43.24 & 29.37 & 0.82 \\
w/ Pick-a-Pic & \underline{54.32} & \underline{30.03} & \underline{0.93} \\
w/ \textbf{UnifiedReward} & \textbf{63.32} & \textbf{32.44} & \textbf{1.05} \\
\midrule
\multicolumn{4}{l}{\textbf{(B) Video Generation DPO (T2V-Turbo)}} \\
{Method} & {Total} & {Quality} & {Semantics} \\
\midrule
Baseline & 80.95 & 82.71 & 73.93 \\
w/ VideoDPO & \underline{81.80} & \underline{83.80} & \underline{73.81} \\
w/ \textbf{UnifiedReward} & \textbf{82.10} & \textbf{84.11} & \textbf{74.06} \\
\bottomrule
\end{tabular}%
}
\vspace{-0.5cm}
\label{tab:gen_dpo_comparison}
\end{table}

%% file: table/sharegptvideo.tex
\begin{table}[tb]
\setlength{\tabcolsep}{6pt}
\centering
\caption{\textbf{Video Understanding Assessment.} We evaluate the performance of our model using different training data configurations.}

\resizebox{\columnwidth}{!}{%
\begin{tabular}{lccccc}
\toprule
\textbf{} & \makecell[c]{OV-7B} & \makecell[c]{w/ Vid.\\ Und.} & \makecell[c]{w/ Vid.\&Img. \\Und.} & \makecell[c]{w/ Vid \\Und.\&Gen.} & \makecell[c]{\textbf{UnifiedReward}}\\ 
\midrule

\textbf{Acc.}            & 48.2 & 74.2 & 76.6 & \underline{78.6} & \textbf{84.0} \\ 
\bottomrule
\end{tabular}%
}
 \vspace{-0.3cm}
\label{tab:sharegpt}
\end{table}

%% file: table/synergy_transfer.tex
\begin{table}[tb]
\centering
\setlength{\tabcolsep}{3.5pt}
\caption{\textbf{Cross-Task Synergy Analysis.} (A) Budget-matched control. (B) Transfer matrix by single-domain training and cross-domain evaluation.}
\resizebox{\columnwidth}{!}{%
\begin{tabular}{lcccc}
\toprule
{Method} & \makecell[c]{{VLReward}\\{Bench}} & \makecell[c]{{ShareGPT}\\{Video}} & \makecell[c]{{GenAI}\\{Image}} & \makecell[c]{{GenAI}\\{Video}} \\
\midrule
\multicolumn{5}{l}{\textbf{(A) Budget-matched control}} \\
Baseline & 29.6 & 48.2 & 53.2 & 50.2 \\
Single-task (native) & 47.4 & 74.2 & 64.0 & 62.4 \\
Single-task (step-matched) & \underline{49.0} & \underline{75.5} & \underline{65.0} & \underline{63.1} \\
\textbf{UnifiedReward} & \textbf{66.1} & \textbf{84.0} & \textbf{70.9} & \textbf{79.3} \\
\midrule
\multicolumn{5}{l}{\textbf{(B) Transfer matrix (single-domain training)}} \\
Baseline & 29.6 & 48.2 & 53.2 & 50.2 \\
Image Understanding-only & \underline{47.4} & 61.5 & 61.8 & 52.5 \\
Image Generation-only & 41.0 & 55.8 & \underline{64.0} & 55.2 \\
Video Understanding-only & 40.2 & \underline{74.2} & 57.5 & 57.8 \\
Video Generation-only & 36.0 & 62.7 & 60.2 & \underline{62.4} \\
\textbf{UnifiedReward} & \textbf{66.1} & \textbf{84.0} & \textbf{70.9} & \textbf{79.3} \\
\bottomrule
\end{tabular}%
}
\vspace{-0.3cm}

\label{tab:synergy_transfer}
\end{table}

%% file: figs/flux_grpo_compare1.tex
\begin{figure*}[t]
    \centering
    \includegraphics[width=1\linewidth]{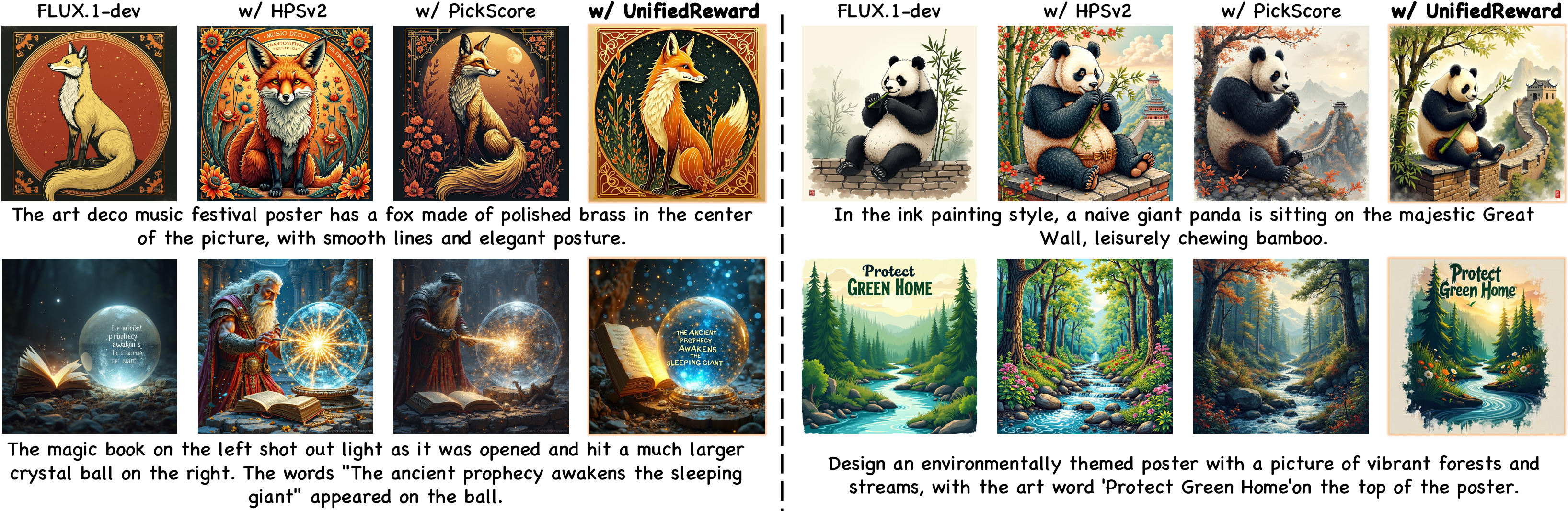}
    \caption{\textbf{Qualitative GRPO Comparison on FLUX.} We compare the visual quality of FLUX.1-dev and GRPO variants optimized with different reward models.}
    \label{fig:flux_grpo_compare}

\end{figure*}

%% file: figs/sdxl_dpo_compare1.tex
\begin{figure}[!t]

    \centering
    \includegraphics[width=0.9\linewidth]{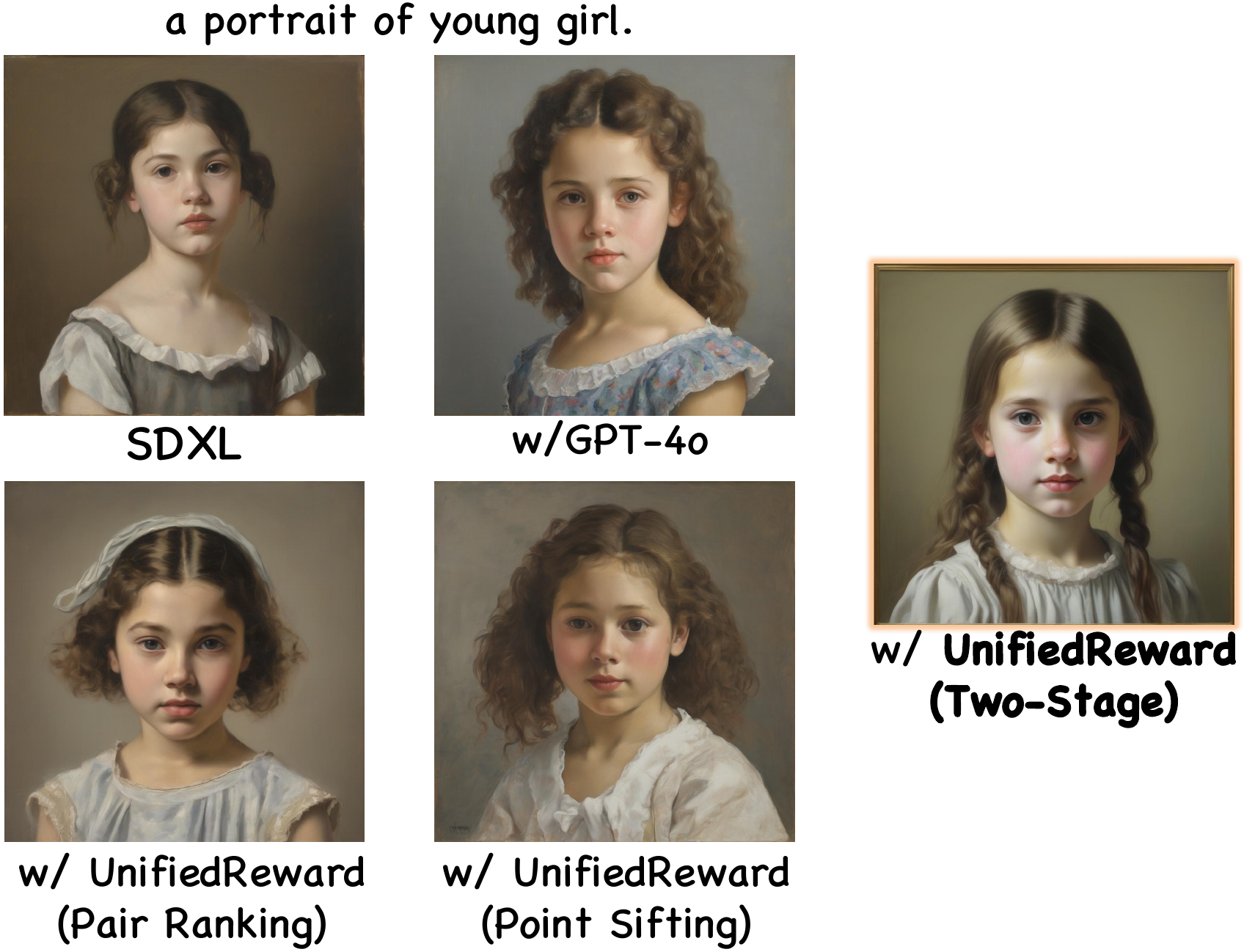}
    \caption{\textbf{Qualitative DPO Comparison on SDXL.} We compare the performance of SDXL, DPO with GPT-4o, UnifiedReward, and UnifiedReward without the pair ranking and point sifting stage.}
    
    \label{fig:sdxl_dpo_compare}
\end{figure}

%% file: table/llava_qwen.tex
\begin{table}[tb]
\setlength{\tabcolsep}{2pt}
\centering
\caption{\textbf{Performance Comparison on Different Backbones.} We compare the performance of \ourmethod trained on LLaVA-OneVision and Qwen2.5-VL.}
\resizebox{\columnwidth}{!}{%
\begin{tabular}{lccccccc}
\toprule
 & \multicolumn{2}{c}{{GenAI-Bench}}& \multicolumn{5}{c}{{VLRewardBench}}\\
\cmidrule(lr){2-3} \cmidrule(lr){4-8}
{UnifiedReward} & Image & Video & General & Hallu. & Reason. & \makecell[c]{{Overall}\\ {Accuracy}} & \makecell[c]{{Macro}\\ {Accuracy}}\\
\midrule

LLaVA-OV-7b       & 70.9 & 77.2    & 60.6 & \textbf{78.4} & 60.5 & 66.1 & 66.5 \\
\hline
Qwen2.5VL-3b       & 68.9 & 78.5    &82.1 & 60.8 & 65.7 & 72.8& 69.5 \\
Qwen2.5VL-7b       & \underline{76.0} & \underline{82.5}    &\underline{84.2} & 68.4 & \underline{73.6} & \underline{77.7}& \underline{75.4} \\
Qwen2.5VL-32b       & \textbf{79.0} & \textbf{85.9}    &\textbf{87.8} & \underline{74.8} & \textbf{75.5} & \textbf{81.5}& \textbf{79.3} \\
\bottomrule
\end{tabular}%
}
\vspace{-0.2cm}
\label{tab:llava_qwen}
\end{table}

%% file: table/imbalance_data.tex
\begin{table}[tb]
\setlength{\tabcolsep}{2pt}
\centering
\caption{\textbf{Performance Under Imbalanced Training-Data Compositions.} The first column denotes the sample amount of \textit{Video Generation : Video Understanding : Image Generation : Image Understanding}; ``K'' indicates thousands of samples.}
\resizebox{\columnwidth}{!}{%
\begin{tabular}{lcccccc}
\toprule
\makecell[l]{Vid.Gen:Und:\\Img.Gen:Und}  & GenAI-Video & ShareGPTVideo & GenAI-Image & VLRewardBench \\
\midrule
\makecell[l]{10:10:10:10K}  &  71.5  & 73.0 & 62.1 &  58.1\\
\midrule
\makecell[l]{10:20:20:20K}   &  71.0  & 74.3 & 64.9 & 60.2  \\
\makecell[l]{10:30:30:30K}   &  69.8  & 75.9  & 66.0 & \underline{61.5}\\
\midrule
\makecell[l]{20:30:30:30K}   &  \underline{72.0}  & \underline{76.2}  & \underline{66.4} & 61.1\\
\makecell[l]{30:30:30:30K}   &  \textbf{72.9}  & \textbf{77.1}  & \textbf{66.9} & \textbf{61.8}\\
\bottomrule
\end{tabular}%
}
\label{tab:imbalance_data}
\end{table}

%% file: table/data_construction_ablation.tex
\begin{table}[tb]
\centering
\setlength{\tabcolsep}{4.5pt}
\caption{\textbf{Preference Construction Analysis.} (A) Preference-signal source comparison. (B) Data-construction strategy ablation. All DPO runs use SDXL and the same optimization budget.}
\resizebox{\columnwidth}{!}{%
\begin{tabular}{lccc}
\toprule
{Method} & PickScore & HPSv2 & ImageReward \\
\midrule
\multicolumn{4}{l}{\textbf{(A) Preference-signal source comparison}} \\
Baseline (SDXL) & 57.82 & 32.61 & 0.84 \\
w/ GPT-4o signal & 59.12 & 32.98 & 0.92 \\
w/ Image-generation-only reward model & \underline{66.12} & \underline{33.46} & \underline{1.04} \\
\textbf{w/ UnifiedReward (ours)} & \textbf{68.28} & \textbf{34.46} & \textbf{1.09} \\
\midrule
\multicolumn{4}{l}{\textbf{(B) Data-construction strategy ablation}} \\
Baseline (SDXL) & 57.82 & 32.61 & 0.84 \\
w/ Random selection & 58.32 & 31.49 & 0.75 \\
w/ Point score only & 60.89 & 32.56 & 0.94 \\
w/ Pair rank only & \underline{62.94} & \underline{33.14} & \underline{1.01} \\
w/ \textbf{Two-stage (ours)} & \textbf{68.28} & \textbf{34.46} & \textbf{1.09} \\
\bottomrule
\end{tabular}%
}
\vspace{-0.2cm}

\label{tab:data_construction_ablation}
\end{table}

%% file: table/grpo_flux.tex
\begin{table}[tb]
\centering
\setlength{\tabcolsep}{9pt}
\caption{\textbf{GRPO Results on FLUX with Different Reward Models.}}
\resizebox{\columnwidth}{!}{%
\begin{tabular}{lcccc}
\toprule
{Method} & {CLIP} & {ImageReward} & {Aesthetic} \\
\midrule
FLUX.1-dev & \underline{34.40} & 1.27 & 6.13 \\
\midrule
w/ HPSv2 & 33.35 & \underline{1.34} & 6.20 \\
w/ PickScore & 33.61 & 1.32 & \underline{6.25} \\
w/ \textbf{UnifiedReward} & \textbf{34.43} & \textbf{1.38} & \textbf{6.31} \\
\bottomrule
\end{tabular}%
}

\label{tab:grpo_flux}
\end{table}

%% file: 5.conclusion.tex
\section{Conclusion}
This paper proposes \ourmethod, the first unified reward model for multimodal understanding and generation assessment, capable of both pair ranking and point scoring, which can be utilized for vision model preference alignment. 
Specifically, we first fine-tune a pre-trained VLM on our constructed large-scale, comprehensive dataset that spans a wide range of visual tasks to develop \ourmethod. This model is then employed to automatically construct high-quality preference pair data from the outputs of vision models through a two-stage filtering process, involving pair ranking and point sifting. These data are subsequently used for model preference alignment via direct preference optimization.
Experimental results demonstrate that joint learning across diverse visual tasks yields significant mutual benefits. By applying our pipeline to both image and video understanding and generation tasks, we achieve substantial improvements in each domain.



%% file: 6.appendix.tex
\clearpage
\appendix

\section{More Implementation Details}

\subsection{Reward Model Baselines}
\noindent\textbf{PickScore} \citep{pickscore} is an image generation assessment model trained over Pick-a-Pic by combining a CLIP-style model with a variant of InstructGPT’s reward model objective. This work employs its checkpoint ``\textit{yuvalkirstain/PickScore\_v1}'' as one of the image generation reward model baselines.

\noindent\textbf{HPSv2} \citep{hps} is an image generation scoring model based on CLIP, fine-tuned on the HPD\_v2 \citep{HPD} dataset. It is capable of predicting human preferences for generated images. We utilize its official code and checkpoint for evaluation.

\noindent\textbf{ImageReward} \citep{xu2023imagereward} is a text-to-image human preference reward model designed to effectively encode human preferences. It is trained based on a systematic annotation pipeline that includes both rating and ranking, collecting 137k expert comparisons. We utilize its official code and checkpoint for evaluation.

\noindent\textbf{LLaVA-Critic} \citep{llava-critic} is designed to assess image understanding performance based on the LLM, enabling pair ranking and point scoring. It is trained on a high-quality critic instruction-following dataset that incorporates diverse evaluation criteria and scenarios. In this work, we employ the ``\textit{lmms-lab/llava-critic-7b}'' model as our baseline for image understanding assessment.

\noindent\textbf{VideoScore} \citep{videoscore} is a video quality assessment model, trained on the VideoFeedback dataset, which contains human-provided multi-aspect scores for 37.6K synthesized videos generated by 11 existing video generative models. We utilize its official code and checkpoint for video quality assessment evaluation.

\noindent\textbf{LiFT} \citep{LiFT} is the first fine-tuning method that leverages human feedback for T2V model alignment. It constructs a Human Rating Annotation dataset, LiFT-HRA, consisting of approximately 20k human annotations, each including a score and its corresponding reason. Based on this dataset, a reward model, LiFT-Critic, is trained to learn a human feedback-based reward function. In this work, we utilize the released code and checkpoint of LiFT-Critic for video generation quality assessment.

\noindent\textbf{VisionReward} \citep{visionreward} is a fine-grained, multi-dimensional reward model designed to capture human preferences in images and videos. It constructs separate human preference datasets for images and videos, and trains corresponding reward models for each. In our work, we utilize its image and video reward models for evaluating image and video generation assessment, respectively.

\noindent\textbf{VideoReward} \citep{videoreward} is a multi-dimensional video reward model trained on a newly proposed 182k-sized human-labeled video generation preference dataset, sourced from 12 video generation models. We utilize its official code and checkpoint for evaluation.

\noindent\textbf{Our UnifiedReward} is based on LLaVA-OneVision-7B (OV-7B) \citep{llava-ov} and trained on our constructed large-scale, comprehensive human feedback dataset, which spans a wide range of visual tasks. Through joint multi-task learning and evaluation, our experimental results demonstrate that this approach fosters a mutually reinforcing effect across tasks. To the best of our knowledge, this is the first unified reward model for multimodal understanding and generation assessment.

\input{figs/supp_qualitative_image1}
\input{figs/supp_qualitative_video1}


\subsection{Evaluation Benchmarks}
\noindent\textbf{Multimodal Understanding}: We evaluate the image and video understanding assessment of \ourmethod on VLRewardBench \citep{vlrewardbench} and ShareGPTVideo \citep{sharegptvideo} (1K samples for testing), respectively. 
\noindent\textbf{Multimodal Generation}: GenAI-Bench \citep{genai-bench} includes both image and video generation reward benchmarks, which are utilized. Besides, we also employ VideoGen-RewardBench \citep{videoreward} for video generation assessment benchmark. 

\subsubsection{Multimodal Understanding}
\textbf{VLRewardBench} \citep{vlrewardbench} is a comprehensive benchmark for assessing image understanding, covering general multimodal queries, visual hallucination detection, and complex reasoning tasks. It consists of 1,250 high-quality examples meticulously designed to evaluate model limitations and challenge their capabilities. During evaluation, we randomly shuffle the order of responses to ensure more robust and reliable assessment results.

\noindent\textbf{ShareGPTVideo} \citep{sharegptvideo} is an open-source, large-scale training dataset comprising 900k captions that cover a diverse range of video content, including temporal dynamics, world knowledge, object attributes, and spatial relationships. It also includes 17k preference data specifically curated for DPO training. In this work, we utilize 16k preference data for reward model training and 1k for video understanding evaluation.

\noindent\subsubsection{Multimodal Generation}
\textbf{GenAI-Bench} \citep{genai-bench} is a reward benchmark for multimodal generative models, designed to assess the ability of MLLMs to evaluate AI-generated content by comparing their judgments with human preferences. It includes benchmarks for image generation, image editing, and video generation. In this work, we utilize the image and video generation parts for generation reward evaluation.

\noindent\textbf{VideoGen-RewardBench} \citep{videoreward} builds upon VideoGen-Eval to establish a fair benchmark for assessing the performance of reward models on modern T2V models. It comprises 26.5k manually constructed video pairs, with annotators evaluating each pair based on Visual Quality, Motion Quality, Text Alignment, and Overall Quality. In this work, we utilize the Overall Quality metric for baseline reward comparison.

We will release all evaluation codes to facilitate community reproduction.

\subsection{DPO Baselines}

\noindent\textbf{LLaVA-Critic} \citep{llava-critic} leverages image-question pairs from LLaVA-RLHF \citep{2023llavarlhf} to construct preference data for OV-7B DPO which is trained for 3 epochs. In this work, for a fair comparison, we also use the image-question pairs from LLaVA-RLHF to construct preference data while keeping all other settings the same.

\noindent\textbf{LLaVA-Houd-DPO} \citep{llava-houd-dpo} utilizes the 17k preference data from the ShareGPTVideo \citep{sharegptvideo} dataset for DPO training. In this work, to ensure a fair comparison, we apply the same dataset for DPO training on LLaVA-Video \citep{llava-video} following its method as the baseline. For our approach, we randomly sample 14k data points from the 17k dataset to construct the DPO training data and then perform DPO on LLaVA-Video. All training parameters and settings are kept identical to maintain fairness in evaluation.

\noindent\textbf{LLaVA-TPO} \citep{li2025temporal} adopts a self-training approach that enables models to distinguish between well-grounded and less accurate temporal responses by leveraging curated preference datasets at two granularities: localized temporal grounding and comprehensive temporal grounding. In this work, since its training dataset has not been open-sourced, we utilize its released checkpoint for comparison.

\noindent\textbf{VideoDPO} \citep{videodpo} is the first video generation DPO method built upon its comprehensive preference scoring system, OmniScore, which evaluates both the visual quality and semantic alignment of generated videos. In this work, we use its released preference dataset for T2V-Turbo \citep{t2v-turbo} DPO as a baseline. For our method, we extract video-caption pairs from its dataset to construct our own preference data for DPO, ensuring a fair evaluation.

\noindent\textbf{Pick-a-Pic} \citep{pickscore} is a large, open dataset of text-to-image prompts paired with real user preferences over generated images. After excluding approximately 12\% of tied pairs, the dataset contains around 851k preference pairs with 58.9k unique prompts. In this work, we directly use this dataset for SDXL-Turbo \citep{sdxl} DPO as a baseline. For our method, we randomly sample 14k captions from this dataset to construct preference data for DPO, ensuring a fair evaluation.


\section{More Qualitative Comparison}
More qualitative results are shown in Figs. \ref{fig:supp_qualitative_image} and \ref{fig:supp_qualitative_video}.

\input{figs/gpt_prompt1}

\section{Societal Impacts}
Our unified reward model for multimodal understanding and generation assessment has the potential to significantly enhance AI applications across various domains. By aligning AI-generated content more closely with human preferences, our work can improve the quality and reliability of vision models, benefiting industries such as digital media, entertainment, education, and accessibility.
For example, one of the key advantages of our approach is its ability to provide a more consistent and interpretable evaluation of generative models. This can lead to better AI-assisted creativity, enabling artists, designers, and content creators to generate higher-quality visuals with greater control. 
While our work brings many benefits, we recognize that reward models, like any AI-driven system, must be carefully designed to ensure fairness and robustness. There is always a risk that biases in the training data could influence model predictions. However, we have taken measures to curate a diverse dataset and will continue refining our approach to mitigate such concerns.
Overall, we believe our work contributes positively to the AI field by providing a more effective and scalable way to align vision models with human preferences. We encourage future research and collaborations to further enhance the fairness, adaptability, and real-world applicability of reward-based AI evaluation.




%% file: figs/supp_qualitative_image1.tex
\begin{figure}[!thb]

    \centering
    \includegraphics[width=1\linewidth]{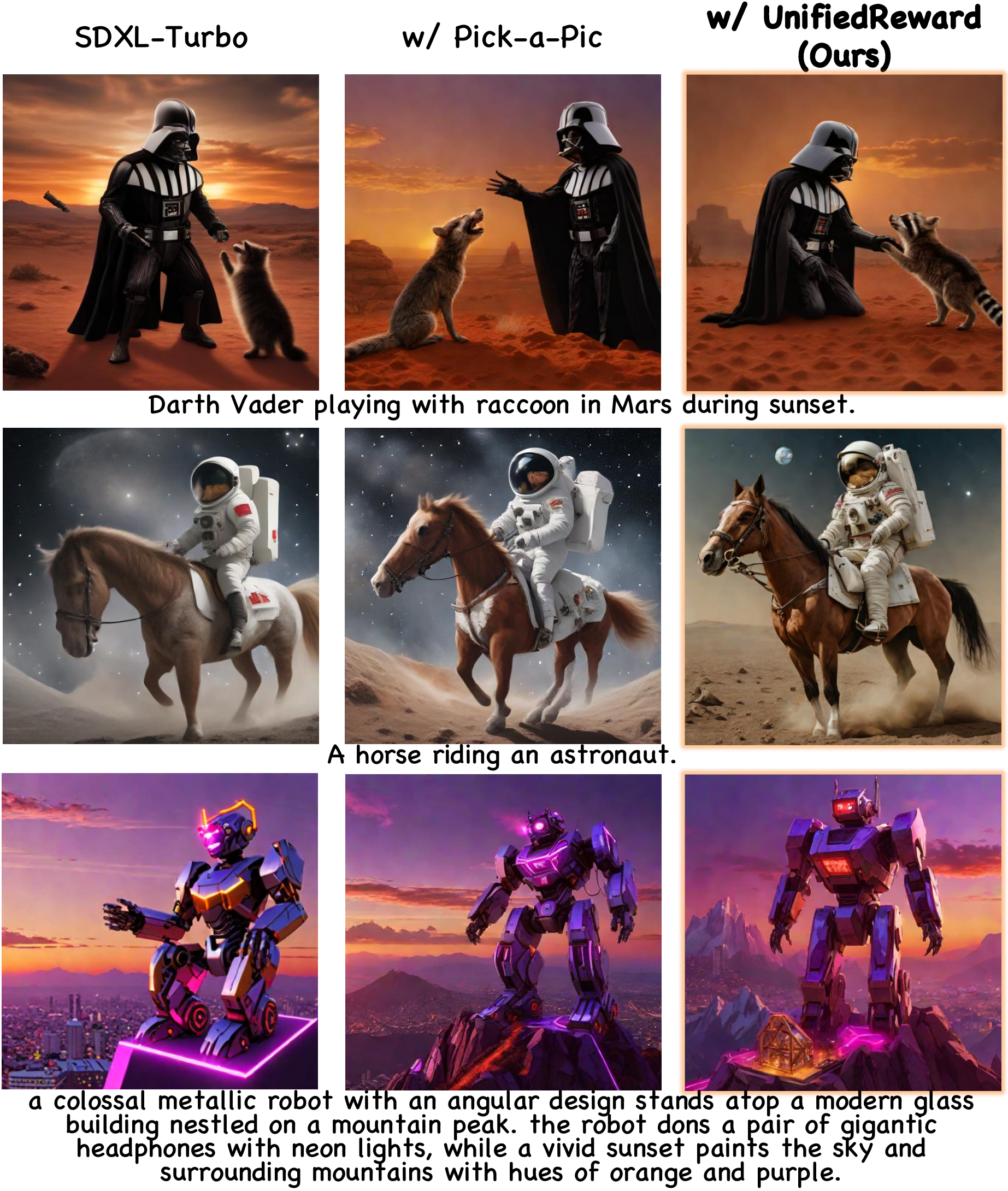}
    \caption{\textbf{More Qualitative DPO Comparison on SDXL-Turbo}. We compare the qualitative performance of the original SDXL-Turbo, DPO trained on Pick-a-Pic dataset, and DPO trained with UnifiedReward.}
    \label{fig:supp_qualitative_image}

\end{figure}

%% file: figs/supp_qualitative_video1.tex
\begin{figure*}[!thb]

    \centering
    \includegraphics[width=1\linewidth]{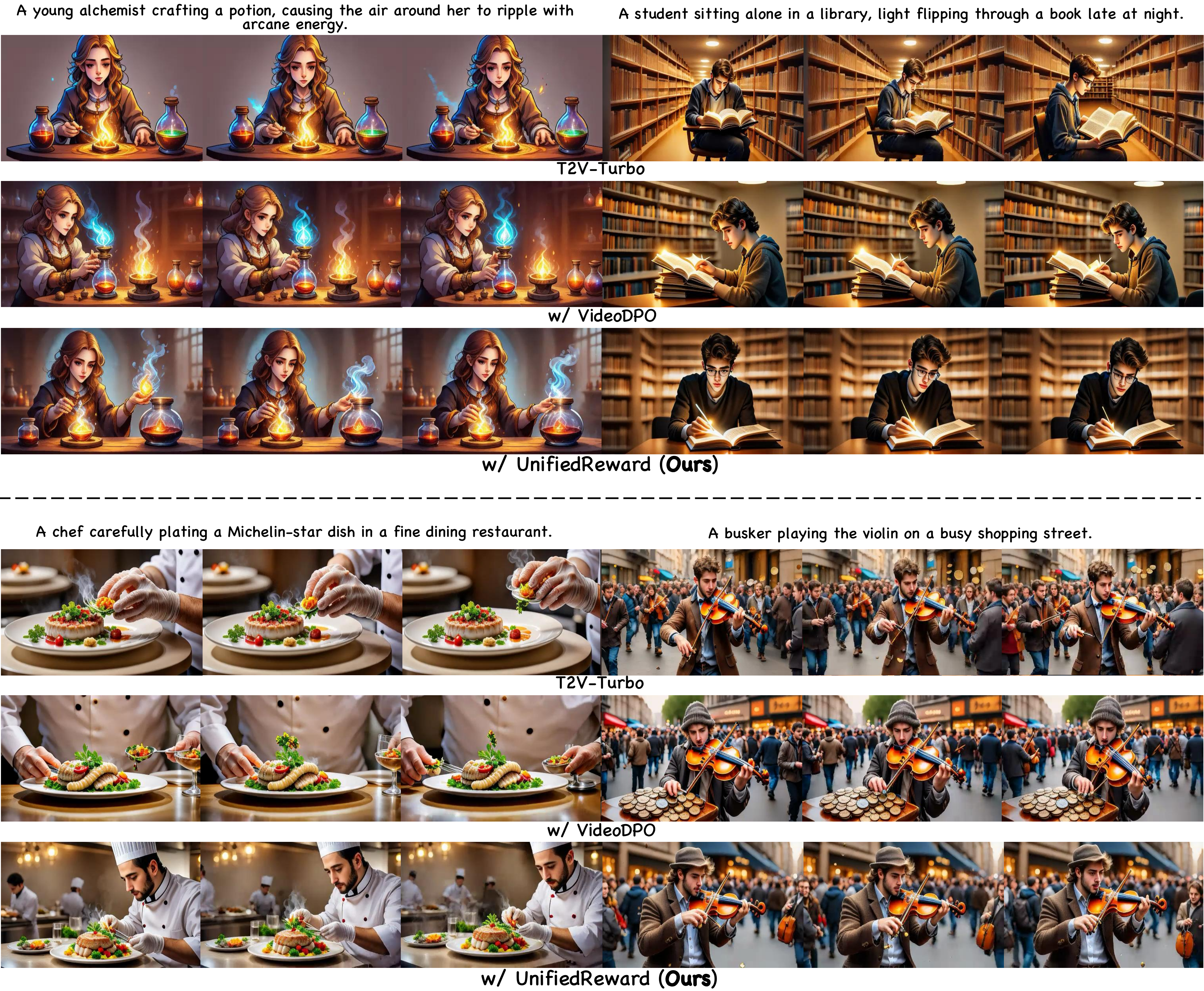}
    \caption{\textbf{More Qualitative DPO Comparison on T2V-Turbo}. We compare the qualitative performance of the original T2V-Turbo, DPO trained with VideoDPO, and DPO trained with UnifiedReward.}
    \label{fig:supp_qualitative_video}

\end{figure*}

%% file: figs/gpt_prompt1.tex
\begin{figure}[tb]

    \centering
    \includegraphics[width=1\linewidth]{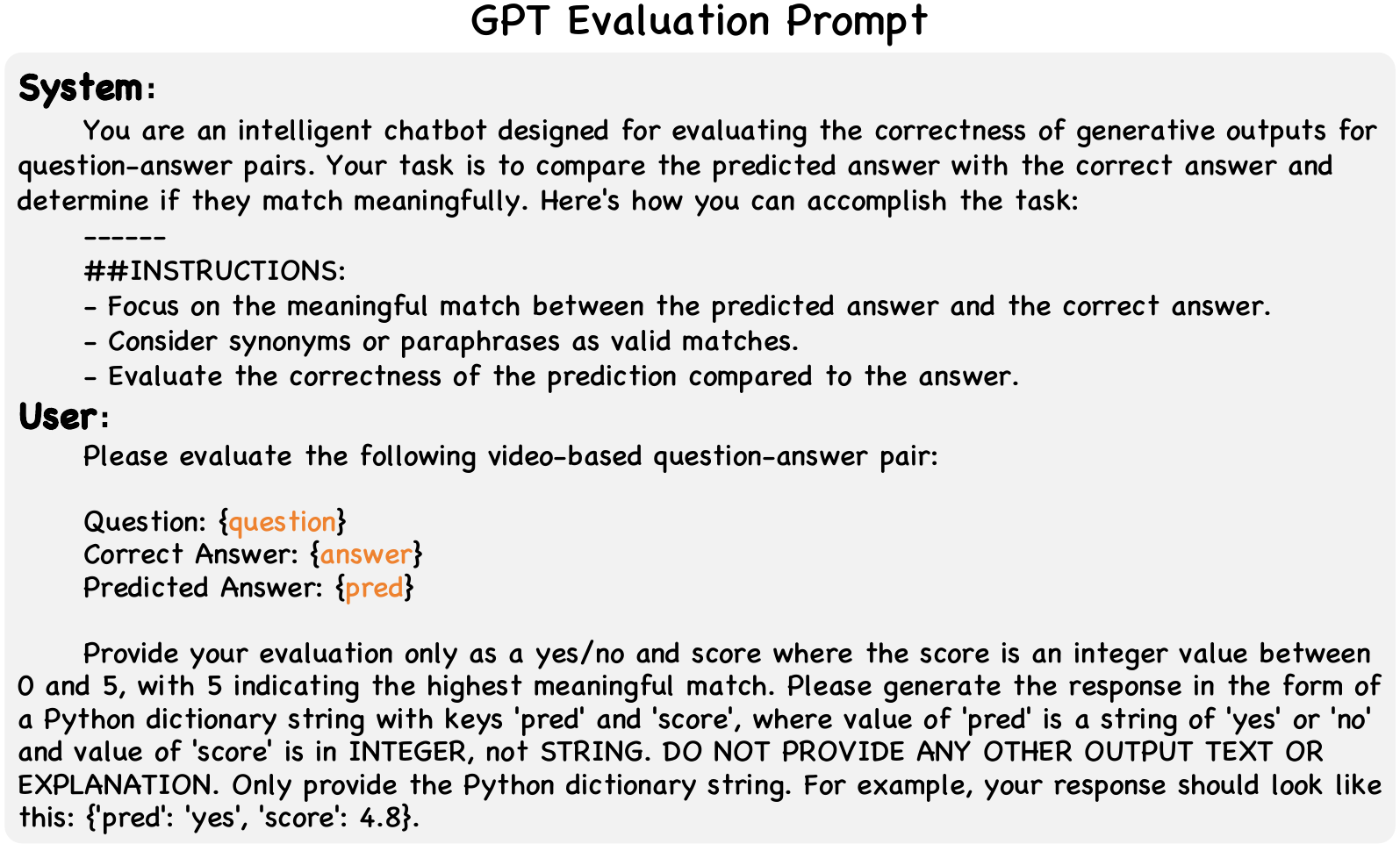}
    \caption{\textbf{GPT Evaluation Prompt}. We use ``gpt-3.5-turbo-1106'' for video understanding evaluation on MSRVTT, MSVD, and TGIF benchmarks.}

    \label{fig:gpt_prompt}

\end{figure}